%% file: root.tex
\documentclass[letterpaper, 10 pt, conference]{ieeeconf}  

\IEEEoverridecommandlockouts                              

\usepackage{hyperref}
\hypersetup{bookmarksopen,bookmarksnumbered,
pdfpagemode=UseOutlines,
colorlinks=true,
linkcolor=blue,
anchorcolor=blue,
citecolor=blue,
filecolor=blue,
menucolor=blue,
urlcolor=blue
}
\usepackage{upgreek}
\usepackage{amsfonts,amssymb}
\usepackage{bm}
\usepackage{bbm}

\usepackage{amsthm} 
\usepackage{thmtools}
\usepackage{mathtools}
\usepackage{xspace}
\usepackage{units}
\usepackage{booktabs} 
\usepackage{tabulary}
\usepackage[usenames,dvipsnames,table]{xcolor} 
\usepackage{diagbox}
\usepackage[ruled,vlined,linesnumbered]{algorithm2e}  
\SetKwComment{Comment}{$\triangleright$\ }{}
\usepackage{ifthen,version}
\usepackage{soul}
\usepackage{csquotes}
\usepackage{microtype}      
\usepackage{lipsum}
\usepackage{tikz}
\let\labelindent\relax
\usepackage{todonotes}
\usepackage{enumitem}
\usepackage[noadjust]{cite}
\usepackage{wrapfig}
\usepackage{graphicx}
\usepackage{array}
\usepackage{subcaption}
\usepackage{nopageno}

\usepackage[normalem]{ulem}
\newcommand\redsout{\bgroup\markoverwith{\textcolor{red}{\rule[0.5ex]{2pt}{0.7pt}}}\ULon}

\newcommand{\xxnote}[3]{}
\ifx\hidenotes\undefined
  \usepackage{color}
  \renewcommand{\xxnote}[3]{\color{#2}{#1: #3}}
\fi

\usepackage{multirow}
\usepackage{pbox}

\newtheoremstyle{hypstyle}
{3pt} 
{3pt} 
{\itshape} 
{} 
{\bfseries} 
{.} 
{.5em} 
{} 

\theoremstyle{hypstyle}

\input{macros/math_operators}

\input{macros/math_defns}
\input{macros/text_shortcuts}
\graphicspath{{figs/}}

\title{\LARGE \bf
Batteries, camera, action!\\
Learning a semantic control space for expressive robot cinematography
\vspace{-2mm}}

\author{Rogerio Bonatti$^{1,2}$, Arthur Bucker$^{3}$, Sebastian Scherer$^{2}$, Mustafa Mukadam$^{1}$ and Jessica Hodgins$^{1}$\\[2mm]
$^{1}$Facebook AI Research, $^{2}$Carnegie Mellon University, $^{3}$University of S\~ao Paulo
\vspace{-3mm}
}

\let\oldtwocolumn\twocolumn
\renewcommand\twocolumn[1][]{%
    \oldtwocolumn[{#1}{
    \begin{center}
           \includegraphics[width=1.0\textwidth]{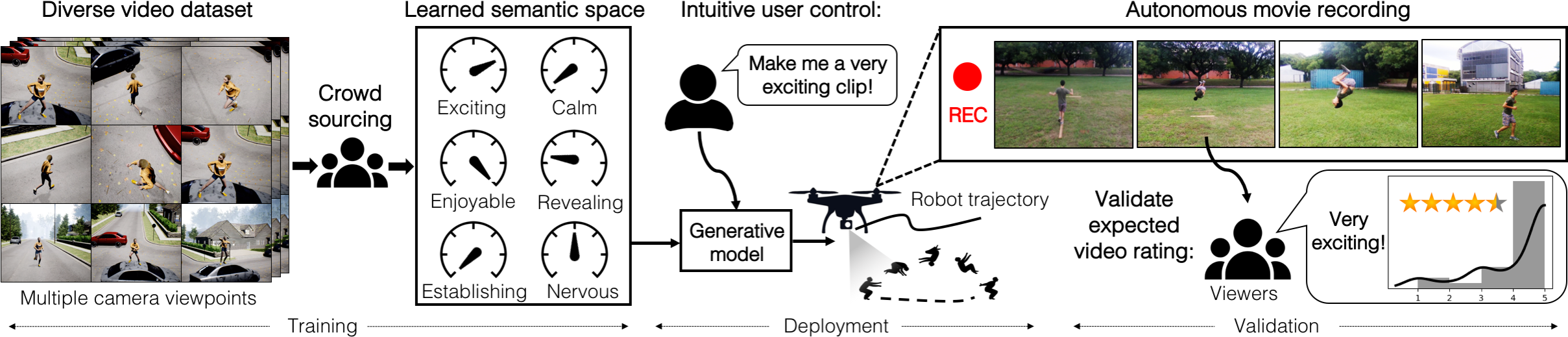}
           \captionof{figure}{\small Our framework uses a diverse set of video clips to learn a semantic descriptor space from crowdsourced ratings. During deployment, a user can intuitively manipulate the robot camera motion with descriptors instead of tuning low-level robot positioning parameters.
    	   }
           \label{fig:main}
        \end{center}
    }]
}

\begin{document}

\maketitle
\thispagestyle{plain}
\pagestyle{plain}


\input{inputs/0_abstract}
\input{inputs/1_intro}
\input{inputs/2_related_work}
\input{inputs/3_perceptual_exp}
\input{inputs/4_control_space}
\input{inputs/5_validation}
\input{inputs/6_discussion}
\section*{Acknowledgments}

Work done while RB interned at Facebook AI Research. RB gratefully acknowledges the support from the Waibel Education Fund. The authors also thank Deepak Gopinath and Jack Urbanek for the help with the crowd-sourcing platform.


\footnotesize{
\bibliographystyle{IEEEtran}
\bibliography{IEEEexample.bib}
}


\end{document}

%% file: macros/math_operators.tex
\newcommand{\real}[0]{\mathbb{R}}

\newcommand{\bbm}{\begin{bmatrix}}
\newcommand{\ebm}{\end{bmatrix}}


%% file: inputs/0_abstract.tex

\begin{abstract}

Aerial vehicles are revolutionizing the way film-makers can capture shots of actors by composing novel aerial and dynamic viewpoints. 
However, despite great advancements in autonomous flight technology, generating expressive camera behaviors is still a challenge and requires non-technical users to edit a large number of unintuitive control parameters.
In this work, we develop a data-driven framework that enables editing of these complex camera positioning parameters in a semantic space (\textit{e.g.} calm, enjoyable, establishing).
First, we generate a database of video clips with a diverse range of shots in a photo-realistic simulator, and use hundreds of participants in a crowd-sourcing framework to obtain scores for a set of semantic descriptors for each clip.
Next, we analyze correlations between descriptors and build a semantic control space based on cinematography guidelines and human perception studies.
Finally, we learn a generative model that can map a set of desired semantic video descriptors into low-level camera trajectory parameters.
We evaluate our system by demonstrating that our model successfully generates shots that are rated by participants as having the expected degrees of expression for each descriptor.
We also show that our models generalize to different scenes in both simulation and real-world experiments.
Data and video found at: \url{https://sites.google.com/view/robotcam}.


\end{abstract}

%% file: inputs/1_intro.tex

\vspace{-1mm}
\section{Introduction}
\vspace{-2mm}

Aerial vehicles have become an important tool for supporting human creativity and expressiveness, fundamentally altering the way both professional and amateur users produce media content for movies, sports and virtual / augmented reality.
Much of the impact of aerial cameras stems from their capability to compose aerial and dynamic viewpoints that are infeasible using traditional devices such as hand-held cameras and dollies \cite{santamarina2018introduction}. 
In addition, recent developments both in industry \cite{mavic2019,skydio2018} and academia \cite{bonatti2020autonomous,jeon2020detection,huang2018act} now allow drones to detect, track and follow objects of interest autonomously while maintaining safety in cluttered environments.

However, a major limitation of today's cinematography platforms is the difficult and unintuitive interface for camera control.
A scene can be captured in myriad ways, and each camera path provides a different visualization of the story \cite{proferes2017film}, causing distinct impressions on the viewers.
Users are not able to directly control the emotional expression of the final footage.
Instead, they need to carefully tune the camera's position, velocities and angles to achieve a desired expression.
An added challenge is the complex interaction between camera parameters, which produces a combinatorial complexity that drives the viewer's perception.
This task is cumbersome to users not only due to the large parameter search space, but also because it requires intimate knowledge of cinematography rules \cite{stanley_2011,bowen2013grammar,arijon1976grammar}, which take years to master.


Our work aims to fill in the gap in existing controls for autonomous cameras.
As seen in Fig.~\ref{fig:main}, our framework enables users to determine the desired shot types based on intuitive and perceptually meaningful parameters (\textit{e.g.} exciting, enjoyable, establishing).
We employ a photo-realistic simulator to generate a database of video clips with a diverse set of shot types, and use crowd-sourced perceptual experiments to obtain semantic scores for each video. We then learn the two-way mapping between low-level shot parameters and semantic descriptors.
We take inspiration from previous works that created semantic control spaces for domains such as cloth animations \cite{sigal2015perceptual}, object and body shapes \cite{yumer2015semantic,streuber2016body}, and robot motion \cite{desai2019geppetto}. Our contributions are threefold:
\\
\textbf{1) Perceptual experiments:}
First, we conduct a series of experiments to determine the minimal perceptually valid step sizes for different shot parameters. We then build a dataset of 200 videos using variations along these units, and use an interactive crowd-sourcing platform to obtain numerical scores for different subjective semantic descriptors;
\\
\textbf{2) Semantic control space:} 
Using the perceptual scores, we learn a regression model that provides an intuitive semantic control space to re-parameterize the desired video expression into low-level shot parameters to guide the aerial camera;
\\
\textbf{3) Experimental validation:} We validate the learned models in a series of experiments in simulation and real-world tests. 
We show that shots generated from the semantic space are rated by participants as having the expected degrees of
expression for each attribute, and that the model generalizes to different actors, activities, and background compositions. 

%% file: inputs/2_related_work.tex
\section{Related Work} 
\label{sec:related_work}

\textbf{Autonomous aerial cinematography:} 
We find multiple lines of work on the use of drones for cinematography. For instance, \cite{roberts2016generating} and \cite{nageli2017real} use optimization methods for navigation on or close to pre-defined trajectories segments.
We also find works that use pre-selected high-level cinematographic guidelines such as distances and angles relative to actors \cite{bowen2013grammar,arijon1976grammar} as an input for trajectory optimization in unscripted scenes among obstacles \cite{bonatti2020autonomous,bonatti2019towards}.
Alternatively, \cite{huang2018through} control the camera based on image projection features. 	

\textbf{Autonomous artistic reasoning:}
Autonomous reasoning about how camera movements can provide artistic value to videos has been a topic of interest in multiple contexts. For instance, \cite{chen2015mimicking} and \cite{chen2017should} train policies to control pan-tilt-zoom cameras in basketball and soccer games by imitating human demonstrations and using video features.
For drone cinematography, \cite{gschwindt2019can} use reinforcement learning to train a policy that switches between four basic shot types to maximize human-provided rewards.
We also find works that aim to imitate different camera motion styles. 
For example, \cite{ashtari2020capturing} use trajectory optimization so that a flying camera can imitate the idiosyncrasies of hand-held camera motions. 
\cite{huang2019one} use supervised learning to predict camera actions for different shot types, and \cite{jiang2020example} build a latent space of shot types out of movie examples, which is coupled with a generative model.


\textbf{Sentiment analysis:}
Unlike the previous works which were focused on \textit{imitating} a particular behavior, sentiment analysis focuses on \textit{understanding} how humans perceive and interpret visual content \cite{cambria2015sentic}.
Most video sentiment models are based on psychology literature and use a combination of the Valence-Arousal-Dominance state model \cite{mehrabian1996pleasure} with semantic descriptors grouped by similarities \cite{russell1980circumplex}.
Several authors explore the relationship between image features and sentiments \cite{baveye2015liris,canini2011affective,machajdik2010affective}, and train models for movie recommendation, summarization, and information retrieval \cite{canini2011affective,hanjalic2005affective}.


\textbf{Learning a perceptual control space:}
Our approach is motivated by previous works that edit processes using semantic or context-specific attributes, as found in the fields of garment simulations \cite{sigal2015perceptual}, 3D shapes \cite{yumer2015semantic,streuber2016body}, and images \cite{laffont2014transient}. Closest to our domain in robotics, \cite{desai2019geppetto} developed a data-driven interface for semantic design of expressive robot motion. All these past works share a common approach: they first build a dataset connecting instances to their intuitive semantic labels (usually via crowd-sourcing), and then use this data to learn a generative model mapping the semantic descriptors back to unintuitive low-level parameters.

Our work is the first to develop a perceptual control space for designing camera trajectories. We allow users without any domain expertise in aerial vehicles or cinematography to capture expressive camera behaviors. 
Unlike previous works that imitate styles, our system does not require the user to search for specific examples of videos they want to indirectly emulate. As we detail next, we show that we can generate any footage by directly selecting its desired semantic descriptors.








%% file: inputs/3_perceptual_exp.tex

\section{Perceptual Experiments} 
\label{sec:perceptual}

\begin{figure}[t]
    \center
    \includegraphics[width=0.45\textwidth]{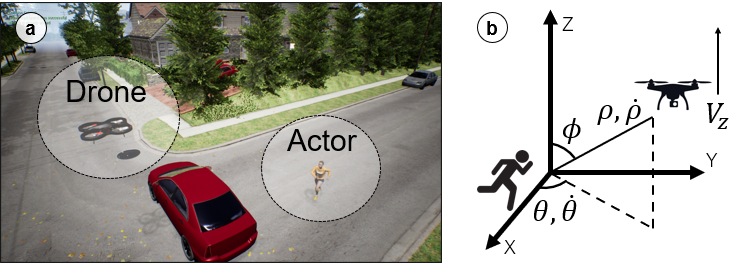}
    \caption{\small Experimental setup: a) Photo-realistic simulator with an animated character and aerial vehicle; b) Shot parameters that define the vehicle's position relative to the actor, in spherical coordinates.
    \vspace{-4.0mm}} 
    \label{fig:sim_and_shot}
\end{figure}

Our overarching goal is to learn a perceptually meaningful space for aerial camera control.
We focus our study on single-actor shots, and assume the existence of sparse sets of obstacles such as the ones found in suburban environments (\textit{e.g.} trees, telephone poles, traffic signs).
All simulated data is recorded using a drone in the photo-realistic environment AirSim~\cite{airsim2017fsr} (Fig.~\ref{fig:sim_and_shot}a) coupled with a custom ROS interface \cite{ros2018}. 
We limit our clips to $15$ second segments, which is a reasonable unit of length for individual shots \cite{mahadani_mahadanii_2015,cutting2010attention}. As seen in the supplementary video, the main scene consists of an animated character running down a street, avoiding a car parked on the road, and finally jumping on top of another vehicle and dancing. We chose this actor motion because it contains both static and dynamic segments, and causes a relatively neutral emotional impression on the viewer.

We employ the base motion planner from \cite{bonatti2019towards}, which uses a trajectory optimization method to avoid collisions and occlusions with the environment. 
Within this framework, a shot is parameterized by the positions and velocities of the drone relative to the actor.
We define a shot using spherical coordinates (Fig.~\ref{fig:sim_and_shot}b): 
$\Omega_{shot} = [\rho, \dot{\rho}, \theta, \dot{\theta}, \phi, V_z ]^T \in \real{}^6$, where
$\rho$ and $\dot{\rho}$ are the distance and velocity towards the actor, $\theta$ and $\dot{\theta}$ are the horizontal angle and angular velocity, $\phi$ is the tilt angle and $V_z$ the vehicle's vertical speed.


\subsection{Minimal Perceptual Units for Shot Parameters}
\label{subsec:perceptual_units}

Our approach requires us to build a mapping between shot parameters and semantic descriptors.
To do so, we must sample the manifold containing all possible shot variations.
Within the cinematography literature \cite{bowen2013grammar,arijon1976grammar} we find canonical sets of values for parameters like the distance to the actor $\rho$ (close-up, medium, and long shots) and tilt angles (variations every $45^{\circ}$).
However, a naive approach for sampling the remaining parameters results in a prohibitively large state space given that some of its parameters such as angle rates are virtually unbounded.
In addition, parameter steps that are too small may render imperceptible changes in the final video, resulting in redundant samples.


To address these issues, we learn a minimally perceptible unit of measure along each dimension in the shot parameter space. A major challenge arises because this is a high-dimensional space, and metrics are not globally consistent (\textit{e.g.} variations in angular velocity will be more perceptible when the camera is closer to the actor). 

\begin{figure}[t]
    \center
    \includegraphics[width=0.49\textwidth]{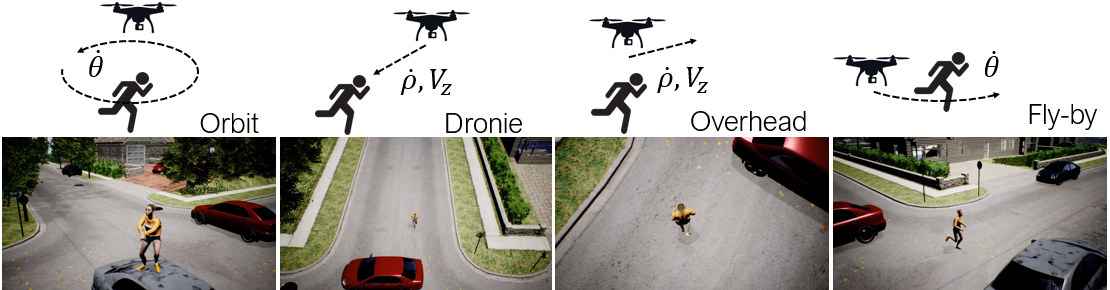}
    \caption{\small Examples of dynamic shot presets used in the study.
    \vspace{-4.0mm}} 
    \label{fig:shots}
\end{figure}

In order to make our experiments tractable we adopt the simplifying assumptions of \cite{sigal2015perceptual}, and evaluate local axis-aligned perceptual steps around a set of meaningful shot preset configurations. 
In Table~\ref{tab:shot_presets} we define six shot presets based on common aerial shot types: the static shots (\textit{Follow 0/1}) maintain a constant relative position between camera and actor, and Fig.~\ref{fig:shots} depicts the dynamic shots.


\begin{table}[ht]
\caption{\small Shot presets for perceptual units study. We only vary the parameters in bold for each preset.
\vspace{-4.0mm}}
\label{tab:shot_presets}
\begin{center}
\begin{tabular}{|c|c|c|c|c|c|c|}
\hline
 & $\rho$  & $\theta$ & $\phi$ & $\dot{\rho}$ & $\dot{\theta}$  & $V_z$  \\ \hline
\backslashbox{Preset}{Unit} & [m]& [$^\circ$]& [$^\circ$]& [m/s] & [$^\circ$/s]& [$^\circ$/s]\\ \hline
Follow 0 & 8  & 0   & \textbf{20} & 0    & 0  & 0 \\ \hline
Follow 1 & 8  & 135 & \textbf{20} & 0    & 0  & 0 \\ \hline
Orbit    & 5  & 0   & \textbf{20} & 0    & \textbf{20} & 0 \\ \hline
Dronie   & 25 & 0   & 45 & \textbf{-0.5} & 0  & \textbf{-0.5} \\ \hline
Overhead & 8  & 180 & \textbf{85} & \textbf{0}    & 0  & \textbf{0} \\ \hline
Fly-by   & 15 & 150 & \textbf{20} & 0    & \textbf{-8} & 0 \\ \hline
\end{tabular}
\end{center}
\end{table}

\vspace{-3.0mm}
For each parameter $\phi$, $\dot{\rho}$, $\dot{\theta}$, $V_z$, we sample approximately ten linearly distributed variations around its preset value over the maximum range.
We only allow changes that keep the shot within its original type.
Table~\ref{tab:shot_presets} highlights the variable parameters in bold.
We create a total of $84$ videos, and for each variation we perform a two-sided t-test analyzing if the resulting video is perceived as the same or different from the preset video (with $p=0.05$ significance).

For this first web survey (WS1) we recruited over $200$ participants using Amazon Mechanical Turk (MTurk) \cite{turkamazon}.
This research received a waiver from our Institutional Review Board, and participants were compensated for their time.
After being approved on a short qualifying task, each participant viewed a total of 12 pairs of videos, one being the preset and the other being either a variation or the preset against itself. Videos were played asynchronously three times (only one video played at a time), and after watching at least once, participants answered: ``\textit{Is the camera perspective the same or different in the two clips shown?}''
Each clip was compared 30 times against its preset.

\begin{figure}[t]
    \center
    \includegraphics[width=0.49\textwidth]{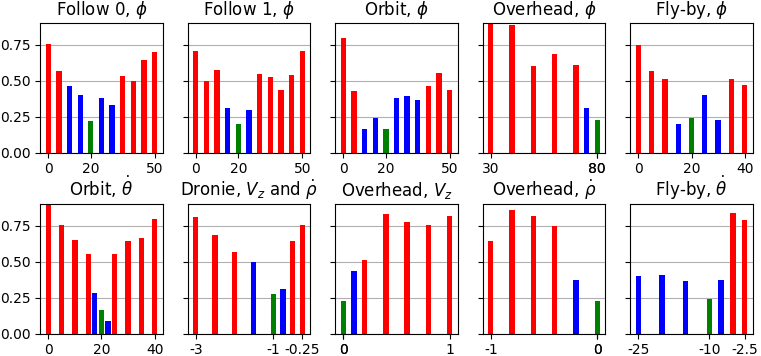}
    \caption{\small Minimal perceptual deltas for each parameter. The preset value is shown in green, red indicates statistically significant perceptual variations ($p=0.05$), and blue are the insignificant variations. Heights display the percentage of videos rated as being different than their preset.
    \vspace{-4.0mm}}
    \label{fig:perceptual_plots}
\end{figure}

Figure~\ref{fig:perceptual_plots} displays our results showing the minimal perceptual units around each preset and parameter. 
As expected, we find that not all noticeable variations are symmetrical around the presets, and deltas are only locally consistent. For example, a change larger than $\Delta\phi = 5^\circ$ in tilt is noticeable for \textit{overhead} shots, while a larger delta $\Delta\phi = 10^\circ$ is required for \textit{follow} shots, where the preset angle is lower.
Humans are surprisingly good at noticing variations in angular speed: deltas as small as $\Delta\dot{\theta} = 2.5^\circ$ were noticeable.


\subsection{Obtaining Semantic Scores for Videos}

Once we are able to generate a diverse dataset of videos with the minimal perceptual units of each shot parameter, the next step is to devise a method to obtain numerical scores for the subjective semantic descriptors representing each clip (our choice of descriptors is detailed in Section~\ref{subsec:descriptor_space}).
To do so, we use \textit{TrueSkill}~\cite{herbrich2007trueskill}, which is a relative ranking algorithm based on pairwise comparisons, similar to the Elo chess-player rating algorithm \cite{elo1978rating}.
Pairwise comparisons have been shown to be more consistent and render more accurate results than absolute scales for subjective scores \cite{baveye2015liris,metallinou2013annotation,yang2010ranking,russell1994ranking}.
We model the score for each semantic descriptor $d_i$ as a Gaussian distribution $d_i = \mathcal{N}(\mu_i, \sigma_i)$, and update both mean and variance of each pair of samples after each comparison. 

\subsection{Building a Semantic Descriptor Space via Crowd-Sourcing}
\label{subsec:descriptor_space}

We initially compiled a list of $15$ semantic descriptors (Table~\ref{tab:descriptor_clusters}) which are commonly used to refer to subjective qualities of images and videos in the cinematography and psychology literature \cite{bowen2013grammar,arijon1976grammar,canini2011affective,baveye2015liris,russell1994ranking,cambria2015sentic}. Because our scenes do not include dialog or soundtrack, we do not include descriptors which are uniquely associated with audio features.

A camera control space consisting of $15$ descriptors is still not an intuitive interface for non-expert users. 
Therefore, in order to reduce the cognitive load on the operator,
our first semantic study targeted reducing the space to a smaller number of descriptors.
To do so, we generated a database of $50$ distinct $15$-second clips using each shot preset's minimal perceptual units. We randomly sampled parameter variations for all axes simultaneously, and allowed positive and negative deltas using multiples of $\{ 0, 0.5, 1.0, 1.5, 2.0\}$ from the perceptual units.

\begin{figure}[ht]
    \center
    \includegraphics[width=0.46\textwidth]{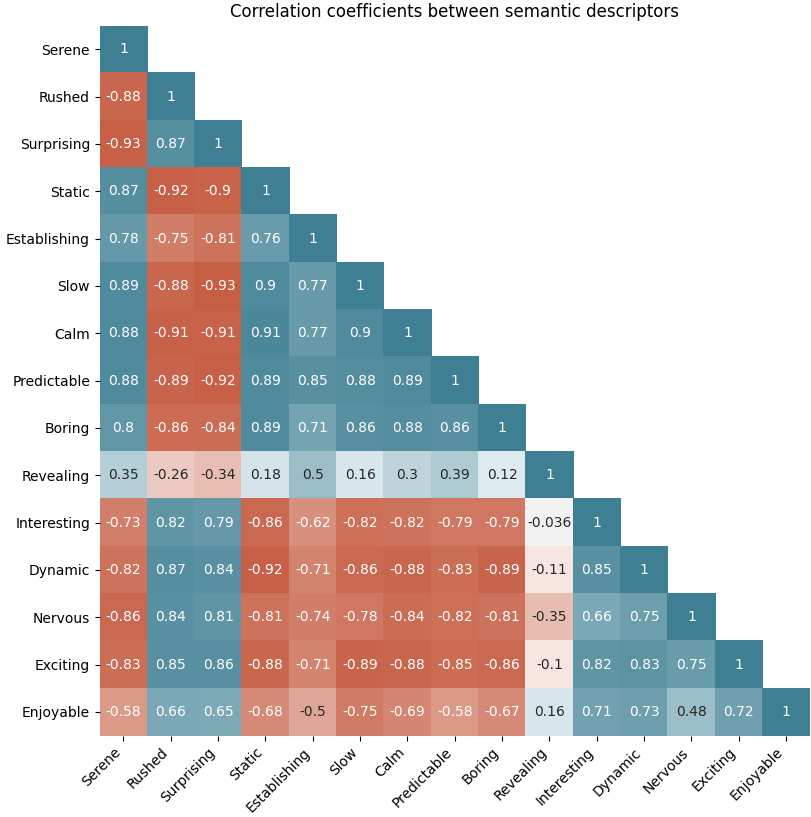}
    \caption{\small Correlations between a set of $15$ semantic descriptors within a diverse dataset of $50$ videos. Data processed from $200$ user surveys. 
    \vspace{-5.0mm}}
    \label{fig:correlations}
\end{figure}

To obtain semantic scores for each shot, we designed a second web survey (WS2) on MTurk.
This time, each user was shown one pair of clips at a time, which played synchronously three times.
After watching the videos at least once, users answered five questions of the type ``\textit{Which video is more \underline{\hspace{.5cm}}?}'', where \underline{\hspace{.5cm}} denotes the different descriptors. Each user watched a total of $12$ video pairs, and passed a qualifying task before the survey. 
We processed answers from a new set of $200$ users using \textit{TrueSkill} to obtain a descriptor vector $d \in \real{}^{15}$ containing the relative scores of each semantic descriptor for each clip.
We analyzed the similarity between all pairs of descriptors using correlation coefficients, as seen in Figure~\ref{fig:correlations}.

\begin{table}[ht]
\caption{\small 7 Clusters of the 15 original descriptors shown within brackets [], and divided by emotion axis and direction. We place no descriptor as representing negative valence, but low values of \textit{interesting} and \textit{enjoyable} can span this emotion.
\vspace{-3.0mm}}
\label{tab:descriptor_clusters}
\begin{center}
\begin{tabular}{|m{3em}|m{11em}|m{5em}|m{5.5em}|}
\hline
Axis & Arousal & Valence & Dominance \\ \hline
Positive & [\textbf{Exciting}, Surprising, Rushed, Dynamic] & [\textbf{Interesting}], [\textbf{Enjoyable}] & [\textbf{Establishing}], [\textbf{Revealing}] \\ \hline
Negative & [\textbf{Calm}, Slow, Predictable, Boring, Serene, Static] & NA & [\textbf{Nervous}] \\ \hline
\end{tabular}
\end{center}
\end{table}

\vspace{-6.0mm}

We found that some groups of descriptors present large positive and negative correlations with one another.
For instance, videos labeled as being very \textit{calm} also present high scores for \textit{slow}, and low scores for \textit{dynamic}. 
We use the Affinity Propagation algorithm \cite{frey2007clustering} to cluster groups of descriptors with high correlation. 
As seen in Table~\ref{tab:descriptor_clusters}, we reduce the dimensionality of the descriptor space by building a total of seven groups of descriptors, with one descriptor representing the group for subsequent studies.
The choice of the number of clusters is ambiguous, and any value between $1$ and $15$ clusters would have been possible by varying the \textit{preference value} in the Affinity Propagation algorithm.
Our choice of seven clusters supported two criteria: (i) it presents a small enough number of parameters that a user can interact within an interface, and (ii) the resulting seven clusters can be grouped to span all axes of the Valence-Arousal-Dominance \cite{mehrabian1996pleasure} space, which is widely used in affective content analysis~\cite{baveye2015liris}. 

\begin{figure}[ht]
    \center
    \includegraphics[width=0.48\textwidth]{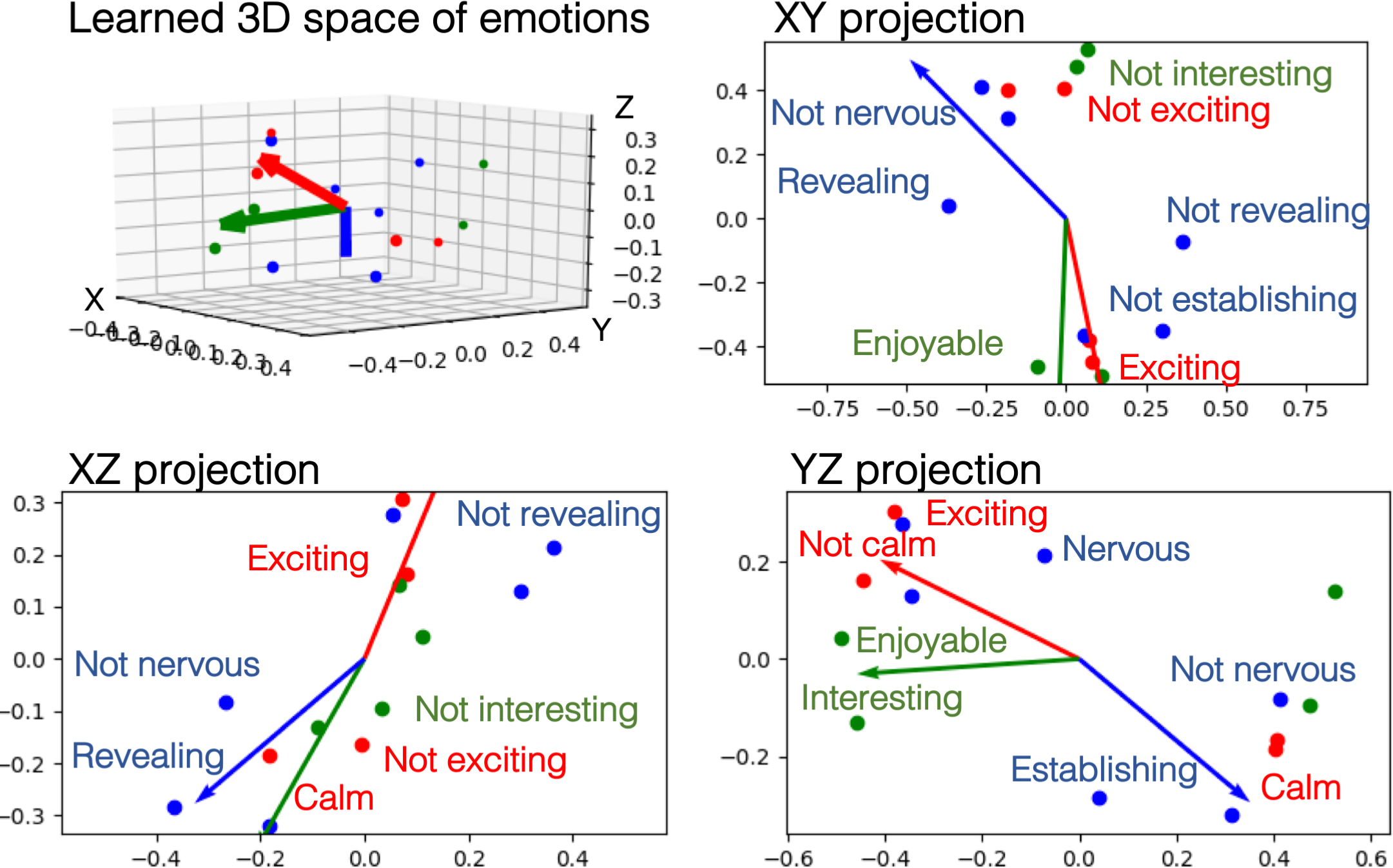}
    \caption{\small Learned 3D space of emotions computed from a survey with $200$ videos and $500$ participants. We display seven semantic descriptors and their negated values (\textit{e.g.} Calm, Not calm), and fit a emotion basis vector along each major emotion axis: Arousal (red), Valence (green) and Dominance (blue).
    \vspace{-4.0mm}}
    \label{fig:3d_space}
\end{figure}

Next, we generated the final dataset for learning the mapping between semantics and shot parameters, now using a larger set of $200$ randomly generated videos sampled around the preset values.
Analogously to WS2, we deployed a third perceptual rating survey (WS3) to compute a $7$-dimensional descriptor vector for each video ($d \in \real{}^{7}$) using answers from a new set of $500$ participants who passed a qualification test.
In order to verify our initial assumption that the resulting clusters are able to span the full space of the three emotions axis, we again calculate a correlation matrix between descriptors. This time we also artificially create negated scores by mirroring each descriptor around its mean value, and then use a multidimensional scaling method \cite{kruskal1964multidimensional} to find the best 3D coordinates that fit our data, treating correlation scores as distances between points. Figure~\ref{fig:3d_space} shows that our experimental data in fact is able to span a 3D affective space, as seen by our fitted Valence-Arousal-Dominance basis vectors.
The vectors are not fully orthogonal, indicating correlations among emotion axes, which is an expected result based on the psychology literature~\cite{russell1994ranking}.

%% file: inputs/4_control_space.tex
\section{Learning a Semantic Control Space} 
\label{sec:semantic_space}

Given the dataset of shot parameters $\Omega_{shot} \in \real{}^6$ and semantic descriptors $d \in \real{}^7$, we wish to learn a function $f: d \rightarrow \Omega_{shot}$ that provides a mapping from a set of desired descriptors to the set of shot parameters that produces a clip with such emotional impression on the viewer (D2P model). We are also interested in the inverse question: given a shot, can we infer its emotional expression as a mapping $f^{-1}: \Omega_{shot} \rightarrow d$ (P2D model)?

\subsection{Training Details}

We explored two approaches to learn such functions: linear regression (LR) and deep neural networks (DNN).
We employ Lasso regression \cite{tibshirani1996regression} for our linear model, as it includes an additional loss to reduce the $L1$ norm of coefficients.
We tested different DNN architectures, and after a series of tests using 5-fold cross validation on our dataset, we settled with a fully connected network with $3$ hidden layers containing $32$, $16$, and $8$ neurons respectively.
We augment the shot parameter vector with $5$ additional variables corresponding to a one-hot encoding of the shot type (follow, orbit, dronie, overhead, fly-by), which is used to process the model's output into one of the canonical drone shot types.
All values are normalized to the $[-1,1]$ range for training.


\subsection{Model Results}

Both LR and DNN presented very similar performance on the D2P model using a dataset of $200$ videos,
with $R_{LR}^2=0.19$ and $R_{DNN}^2=0.22$. Given this similarity and the fact that LR allows us to interpret the values of its coefficients more easily, the rest of our analysis here uses linear models. 
We note, however, that we trained both models with a relatively small quantity of examples due to the high cost of obtaining labeled data. With more data, a DNN model would very likely achieve superior performance.

\begin{figure}[ht]
    \center
    \includegraphics[width=0.45\textwidth]{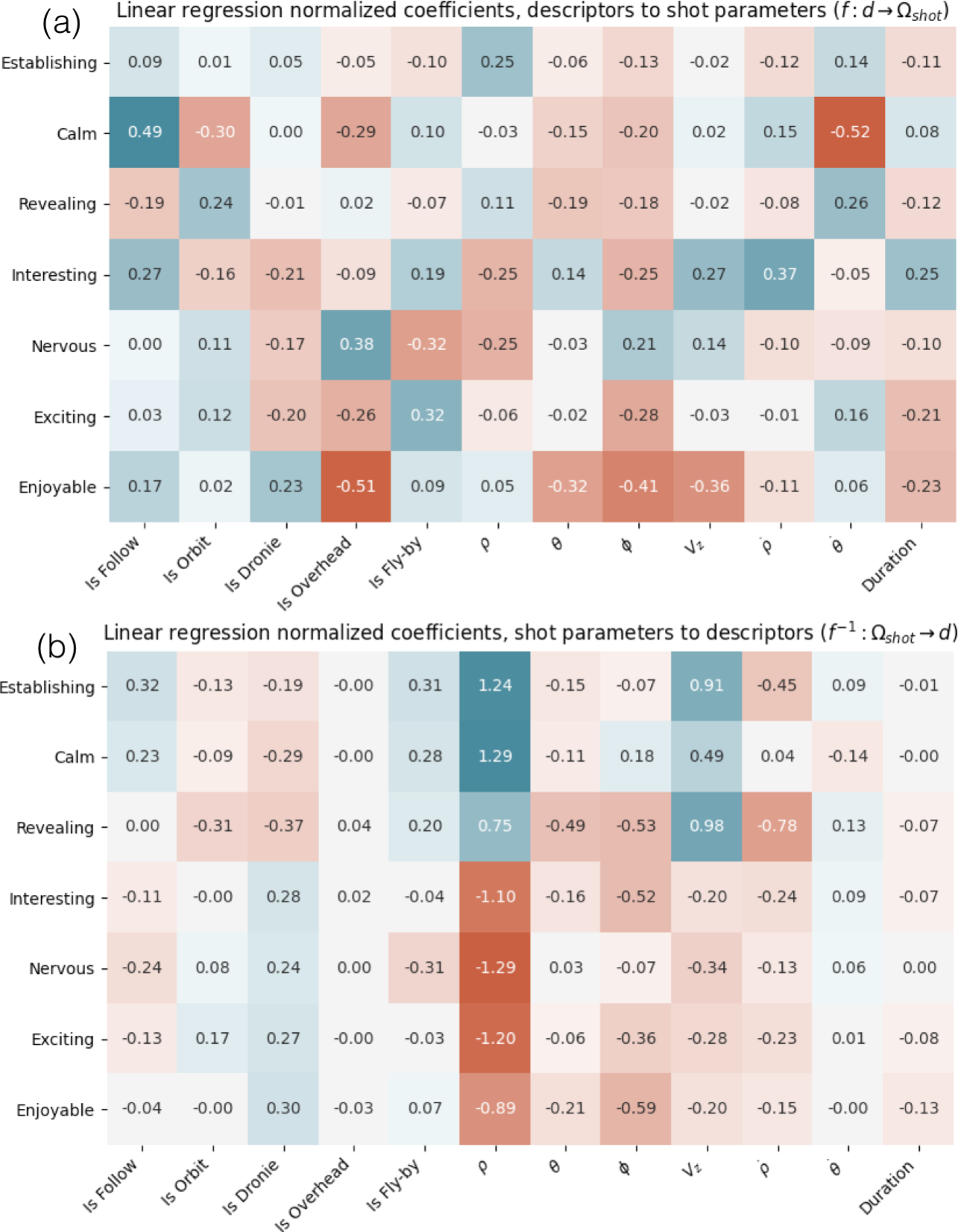}
    \caption{\small Normalized linear model coefficients: a) mapping from semantic descriptors to shot parameters (D2P model), and b) mapping from shot parameters to semantic descriptors (P2D model).
    \vspace{-1.0mm}}
    \label{fig:lin_coeffs}
\end{figure}

Figure~\ref{fig:lin_coeffs}a shows the normalized coefficients for the D2P model.
We notice intuitive relationships between parameters which were learned entirely from user evaluations. For instance, \textit{establishing} videos are linked to larger distances $\rho$ to the actor, matching the cinematographic concept of establishing shots.
Other strong relationships we identify are of \textit{calm} videos having smaller angular velocities $\dot{\theta}$, and enjoyable clips tending to veer towards lower tilt angles $\phi$, away from overhead shots.

We also investigate the P2D i.e. inverse model, which maps shot parameters to descriptor values. Figure~\ref{fig:lin_coeffs}b shows the relationships between variables. The distance between camera and actor causes the largest impact on viewer perception. Other shot parameters have a selective influence on descriptors. For instance, larger tilt angles $\phi$ produce less revealing and less interesting videos, but have less influence on how calm a video is.


\begin{figure}[t]
    \center
    \includegraphics[width=0.45\textwidth]{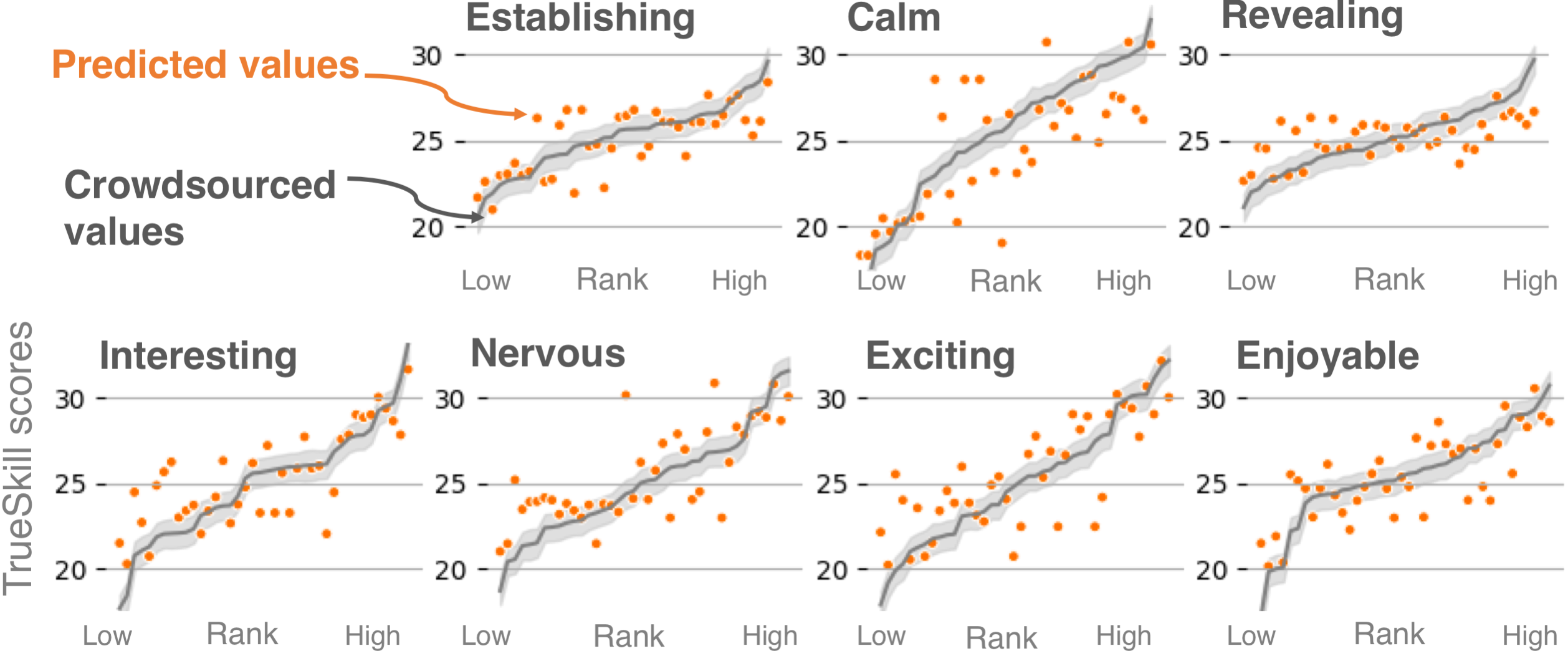}
    \caption{\small Comparison of semantic scores predicted from shot parameters using the learned model (orange) against the original crowdsourced values (gray). Shaded gray area displays TrueSkill score uncertainty. Videos are ordered in ascending order of scores.
    \vspace{-5.0mm}}
    \label{fig:p2d_scores}
\end{figure}

\begin{figure*}[ht]
    \center
    \includegraphics[width=0.95\textwidth]{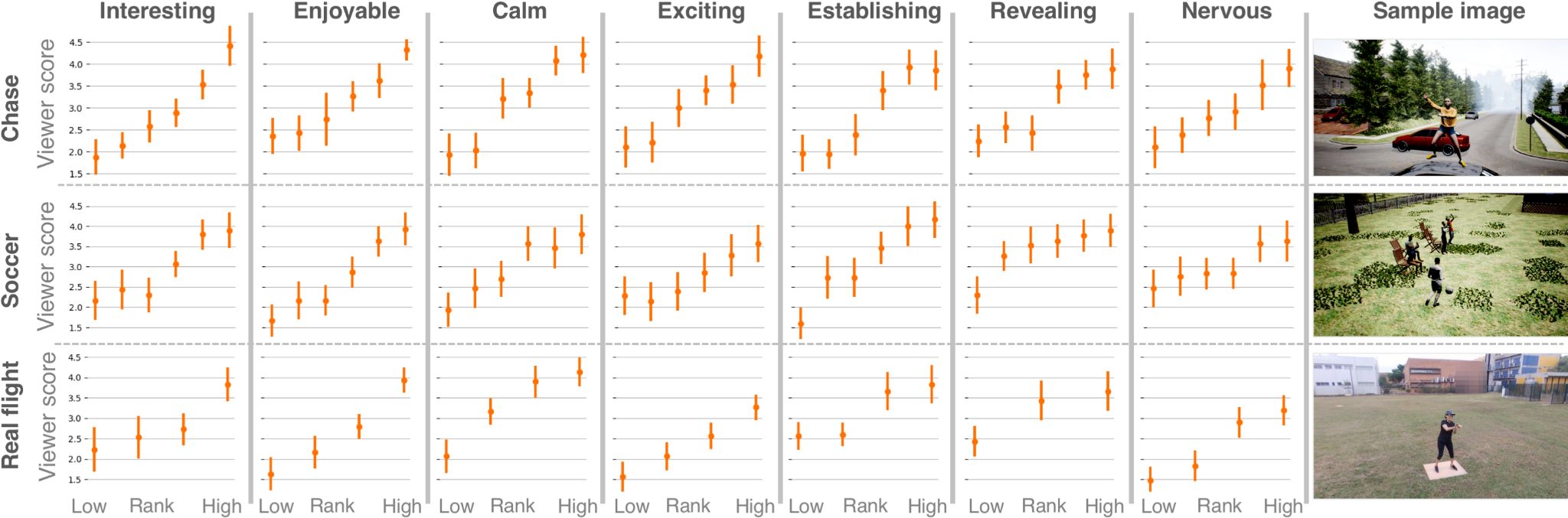}
    \caption{\small User study results validating generative shot model for different simulated and real-world scenes. For each descriptor we generated videos ranging from low to high desired scores, and averaged the perception score from $15$ users on a 5-point absolute Likert scale. Perceived scores match the desired ascending ranking.
    \vspace{-4.0mm}}
    \label{fig:likert_main}
\end{figure*}

\begin{figure}[htb]
    \center
    \includegraphics[width=0.45\textwidth]{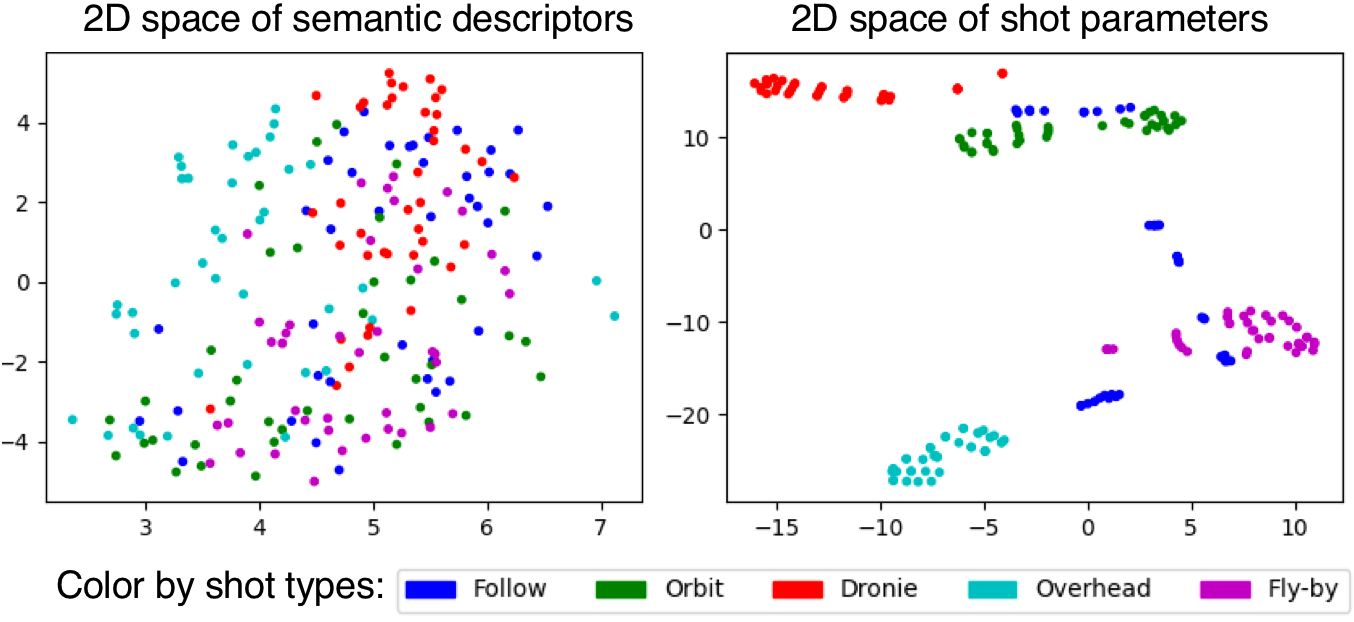}
    \caption{\small Reduction of semantic descriptor space and shot parameter space into 2D compressions using T-SNE. Colors indicate shot types.
    \vspace{-5.0mm}}
    \label{fig:tsne_shots}
\end{figure}

%% file: inputs/5_validation.tex
\section{Experimental Validation} 
\label{sec:validation}

\subsection{Semantic Control Space}

We validated the quality of the learned semantic control space in a series of user experiments. 
Using the D2P model, we generated a set of videos per semantic descriptor by linearly sampling increasingly low to high desired expression values over the range of $[-2, 2]$ standard deviations.
The choice of values for the descriptor vector is not entirely arbitrary, as several of the attributes present positive and negative correlations. For instance, if the user is only interested about the value of a single descriptor and sets a high score for \textit{exciting}, the \textit{calm} score must naturally go down. Therefore, we treat the descriptor vector $d$ as a multivariate normally distributed variable $d = [d_1, d_2]^T$, where $d_2 \in \real{}^n$ are the $n$ user-specified scores. 
We then calculate the remaining scores $d_1$ using the means and covariances between groups of variables:
$d_1 = \mu_1 + \Sigma_{12} \Sigma_{22}^{-1} (d_2 - \mu_2)$.

In contrast to our previous experiments, here we created a new survey on MTurk (WS4) where users viewed each clip individually, and scored them using a 5-point Likert scale ($1$ being \textit{Not \underline{\hspace{.5cm}} at all}, and $5$ being \textit{Extremely \underline{\hspace{.5cm}}}).
We averaged scores from $15$ users per clip. An absolute scale is preferred for this study because our goal is to understand how average users perceive the resulting clips, and not just to create a relative video ranking.

As seen in Figure~\ref{fig:likert_main}, we test our model in two simulation environments using $42$ videos ($6$ per descriptor axis). The first environment is identical to the one employed for learning the models, and the second one tests how well the D2P model generalizes to a new actor type (soccer player dribbling a ball), scene background (grass field with an audience cheering), and shot duration ($20$s, against $15$s earlier). 

In addition, we also validate our model in real-world experiments, collecting a total of $28$ clips ($4$ per descriptor axis).
We use a Parrot Bebop 2 drone and control it with the real-time autonomy stack developed by \cite{bonatti2020autonomous} to visually localize the actor and plan smooth aircraft motions. We collected additional shots (Fig.~\ref{fig:real_life}) shown in the supplementary video.

\begin{figure}[tb]
    \center
    \includegraphics[width=0.47\textwidth]{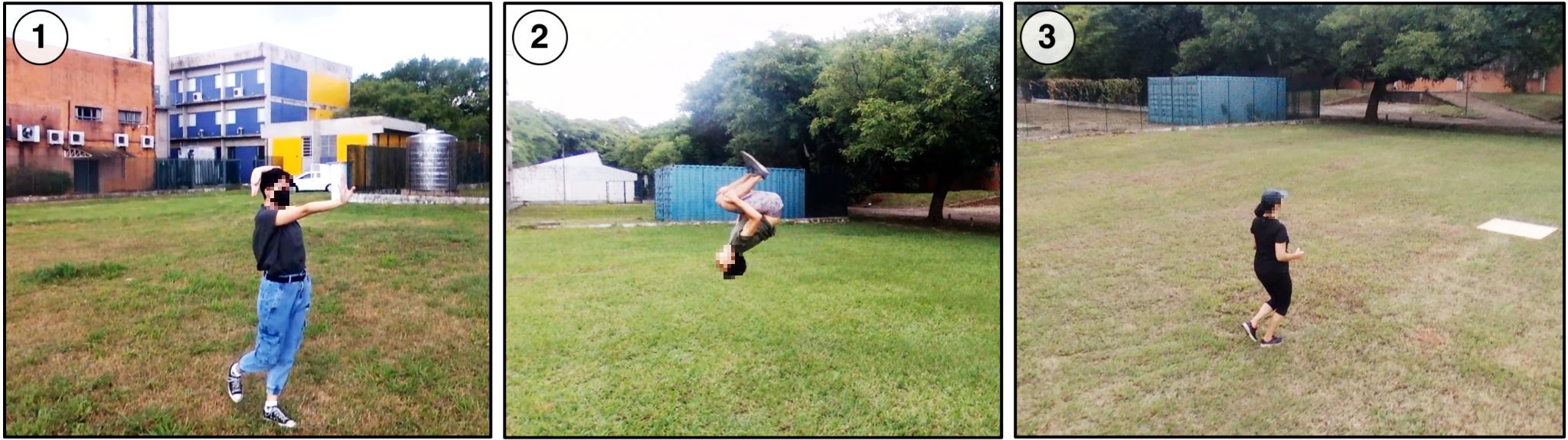}
    \caption{\small Examples of shots taken with our autonomous drone setup: 1) Dance moves; 2) Parkour; 3) Running scenes. The full sequences are shown in the supplementary video.
    \vspace{-6.0mm}}
    \label{fig:real_life}
\end{figure}


Figure~\ref{fig:likert_main} shows that our D2P model is able to successfully generate shots that are rated by participants as
having the expected degrees of expression for each descriptor. Furthermore, the model generalizes well to other simulated scenes and to real-world footages, which strongly suggests that our semantic control space is not overly attached to specific features of the training environment or to a single set of actor motions.

\subsection{Semantic Shot Classification and Latent Space Analysis}

We also validate the quality of the P2D model on a held-out test set with $20$\% of the $200$ videos.
Fig.~\ref{fig:p2d_scores} shows the quality of the model's predictions per emotion ($R^2 = 0.69$). 

Figure~\ref{fig:tsne_shots} provides further insights into the mappings between shot parameters and descriptors. 
We use T-SNE~\cite{maaten2008visualizing} to visualize a 2D projection of higher-dimensional spaces, and color the data by shot types.
Based on the 2D visualization and on the different $R^2$ score values, we argue that learning a model that maps shot parameters to descriptors is an easier task, as it requires learning a surjective function (many-to-one). The inverse mapping is harder since one point in the descriptor space can correspond to multiple set of shot parameters (one-to-many).



%% file: inputs/6_discussion.tex

\section{Conclusion and Discussion}
\label{sec:discussion}

In this paper, we present a semantic control space for aerial cinematography that allows users to guide camera motions using intuitive controls instead of selecting low-level shot parameters.
We use crowdsourcing with hundreds of users to learn the mapping between shot parameters and semantic descriptors in a diverse dataset of $200$ videos.
We validate our semantic generative shot model in multiple experiments, and show that it successfully generalizes to new scenes in simulation and in real-world experiments with an
aerial robot.
Additionally, we provide insights into the space of emotions our model can represent, and investigate the relationship between shot parameters and descriptors to understand what our model is effectively learning.

Our framework targets non-technical users, and can generate shot parameters directly from a semantic vector. 
However, expert users can easily adapt it to gain more control over the model's outcome. 
For example, one can learn separate generative models for individual shot types and gain more control over the system's inputs / outputs.

We find multiple directions for future work in the area.
For instance, we are interested in employing a larger set of parameters to control the shots, such as lens zoom \cite{pueyo2020cinemairsim}, and potentially even soundtracks.
In addition, we would like to extend our framework to accept environment features into the generative model.
Another limitation we face is regarding the shot time horizon: currently our algorithm generates a single shot at a time. However, there is great potential in developing algorithms that can reason over longer durations, and infer emotional expression over sequences of shots.